# Android Kontrollü IP Kameralı Gezgin Robot Tasarımı

# Android Controlled Mobile Robot Design with IP Camera

*Emre DEMİR*
*Bilgisayar Mühendisliği Bölümü*
*İskenderun Teknik Üniversitesi*
*İskenderun, Hatay*
emredemir1995@gmail.com

*Ahmet GÖKÇEN*, *Yakup KUTLU*
*Bilgisayar Mühendisliği Bölümü*
*İskenderun Teknik Üniversitesi*
*İskenderun,Hatay*
{ahmet.gokcen, yakup.kutlu}@iste.edu.tr

*Özet*—Bu çalışmada Arduino kart temelli gezgin robot tasarımı gerçekleştirilmiştir. Bu robot, güvenlik robotu, yardımcı robot veya kontrol robotu olarak görev yapabilecek şekilde tasarlanmıştır. Tasarlanan robotun iki modlu çalışmasına uygun bir algoritma geliştirilmiştir. Birinci çalışma modu otonom moddur. Bu mod da robot, etrafına yerleştirilen ultrasonik sensörler yardımıyla etrafını algılamakta, ayrıca encoder yardımıyla geçtiği yerlerin kaydını tutmaktadır. Bu sayede herhangi bir yere çarpmadan ve geçtiği yerden bir daha geçmeden gezebilmekte ve gezerken üzerindeki IP kamera sayesinde çevresinin görüntüsünü ve ayrıca bilgilendirme ve uyarı modulü ile hastanın nabız ve sıcaklık durumu gibi bilgileri düzenli olarak takip ederek olası anormalliklerde uyarı yapabilmektedir. Hastanın yanında taşıyabildiği acil durum butonu sayesinde acil durumlarda ilgili yerlere bilgi gönderilmektedir. İlaç kullanım zamanları ayarlanarak ilaç zamanı geldiğinde uyarı vererek hatırlatma yapabilmektedir. İkinci mod ise manuel moddur. Bu modda kullanıcı Android işletim sistemine sahip bir cihazla robotu istediği yöne hareket ettirebilmektedir. Böylelikle kullanıcı robotun yanında olmasa da IP kamera görüntsü ile robotu istediği noktaya hareket ettirebilmektedir.

*Anahtar Kelimeler*— Arduino, Android, Gezgin Robot

*Abstract*—In this study Arduino card based mobile robot design was realized. This robot can serve as a security robot, an auxiliary robot or a control robot. The designed robot has two operation modes. The first operating mode is autonomous mode. In this mode, the robot detects the surroundings with the help of ultrasonic sensors placed around it, and keeps track of the places it passes by using the encoder. It is able to navigate without hitting any place and passing from where it passes, and it transmits the patient's pulse and temperature condition to the user by other systems installed on it. Also the IP camera sends the scene on the screen. The emergency button to be placed next to the patient sends information to the user in emergency situations. If the abnormality is detected in the temperature and pulse again, the user gives a message. When the pre-recorded drug use times come, the system can alert the patient. The second mode is manual mode. In this mode, the user can move the desired direction of the robot with the Android operating system. In addition, all data received in autonomous mode can be sent to the user. Thus, the user can control the mobile robot with the camera image even if it is not in the vicinity of the robot.

I. GİRİŞ

Robotlar sensörler ile çevresini algılayan, algıladıklarını yorumlayan, yapay zekâ tekniklerini kullanarak karar veren ve karar sonucuna göre hareket eden aygıtlar olarak otonom veya önceden programlanmış görevleri yerine getirebilen elektro-mekanik cihazlardır.

Günümüzde robotların en büyük kullanım alanı endüstriyel üretimdir. Özellikle otomotiv endüstrisinde çok sayıda robot kullanılır. Bunların çoğu robot kol şeklindeki robotlardır. Bunlar parçaları monte eden, birleştiren, kaynak ve boya yapan robotlardır. Evlerde de robot kullanımı giderek artmaktadır. Evdeki temizlik hizmetlerini yürüten, servis yapabilen robotlar hali hazırda mevcuttur. Yemek yapan robotlar ile ilgili çalışmaların yapıldığı bilinmektedir. Bunun yanında son zamanlarda robotlar, rehabilitasyon merkezlerinde fizyoterapi aktivitelerinde de önemli rol oynamaktadır. Burada otomatik eğitim sistemleri oluşturarak istenilen bölgenin tedavisinde daha iyi gelişim göstermesi sağlanmaktadır [1]. Bu alanda literatürde birçok çalışma mevcuttur. Yine bir diğer çalışmada robotların sosyal etkileşim yoluyla kişilere yardım etmesi sağlanmaktadır [2].

Robotların sağlık / bakım ve sosyal bakım uygulamalarına yönelik çalışmalarda mevcuttur[3]. Bunların dışında robotlar askeri alanlarda, güvenlik güçlerine destek olmada, bomba imha etmede insana yardımcı olmak amacıyla kullanılmaktadır [4].

Türkiye İstatistik Kurumu verilerine göre 2015 yılı Türkiye nüfusunun %28.86'sı 45 yaş ve üstüdür. Bunların %8.6'sı ise 65 yaş üstü kişilerden oluşmaktadır[5]. Ayrıca, kentleşme ve sağlıksız yaşam tarzı sebebiyle insanlar şeker hastalığı, KOAH, obez vb. gibi kronik hastalıkların görülme sıklığı daha da artmaktadır [6].

Bu kronik hastaların yaklaşık %50 sine tavsiye edilen tedaviyi yaptırabilmek büyük sorun teşkil etmektedir. Özellikle de yaşlılar için bu problem daha zordur [7].Bunun yanında çalışan ailelerin en büyük sorunu evde bıraktıkları aile bireylerinin, yaşlı anne baba veya bebeklerinin ve bebek bakıcılarının durumlarını sürekli kontrol etme istekleridir. Bu bağlamda güvenlik kameralar mevcuttur ama her odaya kamera takılması gerekliliği ortaya çıkmaktadır.

Bu çalışmada; ekonomik ve çok amaçlı bir gezgin robot tasarımı gerçekleştirilmiş ve uygulaması yapılmıştır. Tasarlanan robot güçlü bir donanıma, iyi bir yazılıma ve bunun yanında hareket, denetim yeteneğine sahiptir [8]. Bu amaçla kullanımı son zamanlarda çok yaygın olan Arduino kart [9] ile oluşturulan gezgin robot, cep telefonu veya tablet ile uzaktan kontrol edilerek hem otonom hem de manuel olarak hareket edebilmektedir. Ayrıca üzerinde bulunan çok yönlü IP kamera, router aracılığıyla Android sistemli cihaza anlık görüntü göndermektedir. Böylece gezgin robot istenilen noktalara hareket ettirilirken bulunduğu ortamdan sürekli olarak görüntü almamızı sağlamaktadır. Bu yolla robot, medikal alanda mobil refakatçi olarak veya evlerinde bakıcı bulunan aileler evin durumunu sürekli kontrol etmelerinin sağlamaktadır.

II. MATERYAL VE YÖNTEM

Çok amaçlı gezgin robotun geliştirilmesinde kullanılan modüler yapı Şekil 1' de gösterilmiştir. Buna göre Merkezi Kontrol Ünitesi modülünde Arduino MEGA kontrol kartı kullanılmıştır. Dalgaların yansıma özelliğinden faydalanılarak üretilen ultrasonik sensörler [10] sensör modülü için düşünülmüştür.

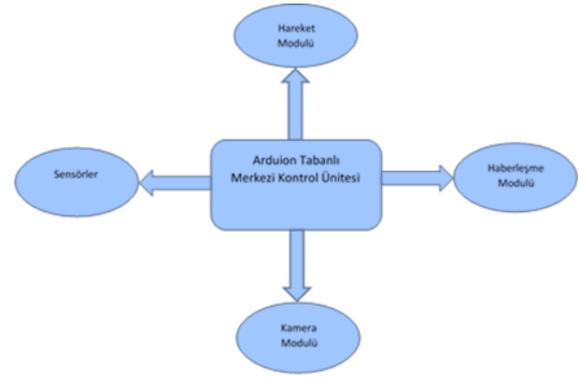

Şekil 1. Çok Amaçlı Gezgin Robot Modüler Yapısı

Robotun etrafına 6 adet HC-SR04 ultrasonik sensör yerleştirilmiştir. Motorlara encoder entegre edilmiştir. Böylece robotun otonom modda çevresini algılayarak bir yere çarpmadan kendi kendine dolaşabilmesi ve bu sayede çevresini öğrenmesi sağlanmaktadır. Nabız, sıcaklık ve acil durum butonu için ayrı bir devre kurulmuştur. Kurulan bu devre robottan bağımsız bir Arduino kartı ile oluşturulmuş, nabız ve sıcaklık sensörleri kullanılmıştır. Bu modül giyilebilir bir tasarım olması sebebiyle sürekli kişinin üzerinde bulundurulmaktadır. Olası acil durum anında üzerindeki buton sayesinde yardım mesajı göndermektedir. Nabız ve sıcaklık için oluşabilecek olağan dışı değerlerde kullanıcıya mesaj olarak gönderilmektedir.

Hareket modülünde 2 adet 6V 500Rpm Redüktörlü Mikro DC Motor kullanılmıştır. Ayrıca robotun dengede durabilmesi için 2 adet serbest tekerden yararlanılmıştır. Haberleşme modülü için Wifi kart kullanılmıştır. Android işletim sistemleri herhangi bir cihaz ile bu modül üzerinden bilgi aktarımı sağlanmakta ve bu sayede robot istenilen noktaya hareket ettirilebilmektedir. IP kamera algılamış olduğu görüntüyü anlık olarak Android cihaza aktarmakta ve kullanıcı uzaktan görüntü erişimine sahip olmaktadır. Görüntü aktarımı her iki modda da (otonom ve manuel) gerçekleşmektedir. Bununla beraber IP kamera 4 yönlü hareket etme yeteneğine sahiptir. Böylelikle robot dolaşırken IP kamera istenilen noktaya çevrilebilmektedir. Güç ünitesi olarak 2 hücreli 4000mAh 7.4V'luk lipo pil kullanılmıştır.

Şekil 2 'de tasarlanan çok amaçlı gezgin robot tasarımı görülmektedir. Şekil 3 'te ise kullanıcının gezgin robot ile bağlantı kurması sağlayan Android arayüzü görülmektedir. Aynı ekranda ilaç saatlerini düzenlenme menüsü ve nabız, sıcaklık gibi değerler görünmektedir.

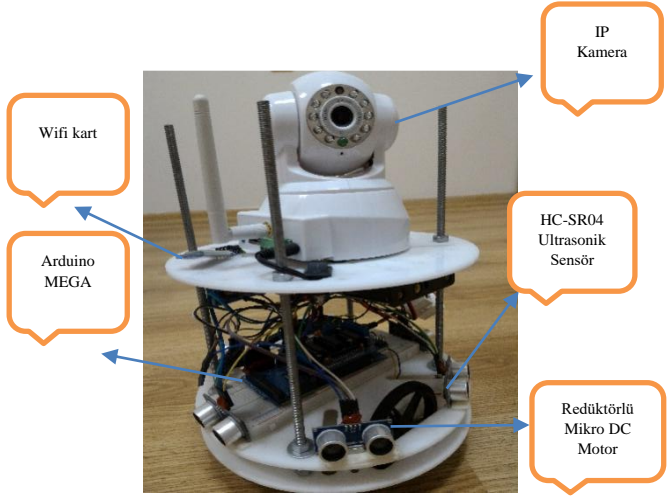

Şekil 2. Çok amaçlı gezgin robot tasarımı

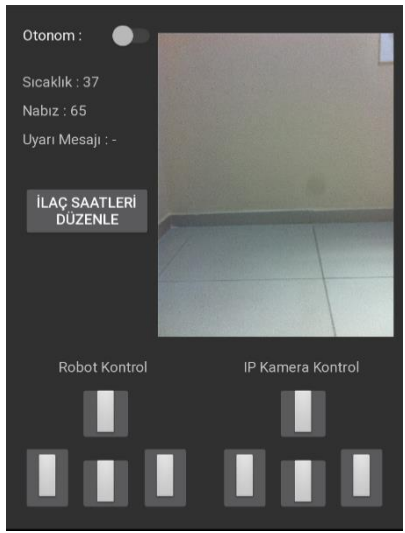

Şekil 3. Android Arayüzü

Çok amaçlı gezgin robotun çalışma algoritma akış diyagramı Şekil 4'te gösterilmiştir.

Robot, Android işletim sistemli herhangi bir cihaz ile çalışabilmektedir. Yazılım açıldığı anda varsayılan mod manuel moddur. Otonom seçeneği aktif edildikten sonra robot ultrasonik sensörler ile ortam taraması yapıp kendiliğinden harekete başlayacaktır. Ultrasonik sensörler yardımıyla hiçbir engele takılmadan hareketine devam edebilmektedir. Bununla birlikte encoderlerden alınan değerler kaydedilerek bir algoritma ile ortamı öğrenmesi sağlanmış ve sürekli aynı yerde dolaşması engellenmiştir. Böylelikle farklı noktaları gezebilmekte ve her noktadan görüntü aktarımı yapabilmektedir. Ayrıca hastanın nabız ve sıcaklık bilgilerini göstermekte anormal bir durum tespit edildiğinde uyarı vermektedir. İlaç hatırlatma modülü ilaç saatlerini hatırlatmaktadır. Hasta acil durum butonuna bastığı takdirde kullanıcıya mesaj gönderilmektedir. İstenirse robot, manuel olarak da hareket edebilmektedir. Bu sayede evde güvenlik robotu, bakıcı kontrol robotu, refakatçi robot olarak görev yapabilmektedir. Kamera istenildiği anda istenilen yöne çevrilebilmekte ve evin içinde istenilen noktadan görüntü alınabilmektedir.

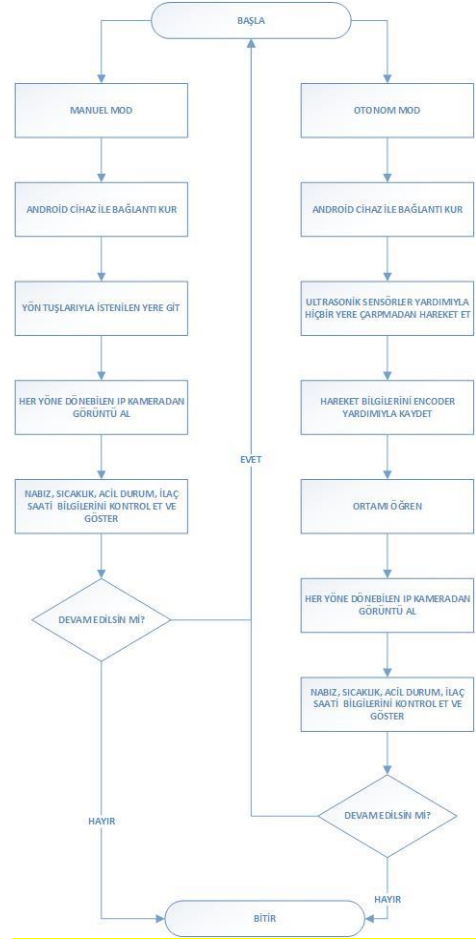

Şekil 4. Akış diyagramı

*A. Ultrasonic Sensörler*

Bir cismin varlığını belirlemek için ultrasonik dalgalar kullanılır. Ultrasonik ses dalgaları çarptığı cisimlerden geriye yansır. Bu yansıma kullanılarak cisimlerin varlığı belirlenebilir. Metaller, ahşap cisimler, sıvılar, camlar, plastik malzemeler ve kâğıt gibi ürünler ultrasonik ses dalgalarının %100'e yakınını geriye yansıtırlar. Bunun yanında pamuk ve yünlü bezler ultrasonik dalgaları emerler. Ultrasonik dalgaların bu yansıma özelliğinden faydalanılarak ultrasonik sensörler üretilmiştir [11-13].

Ultrasonik ses dalgalarının hareketi aşağıdaki formülle açıklanır :

$$C = 331{,}5 + 0{,}607t \ (m/s) \quad (1)$$

Formülden görüleceği gibi dalgaların hareketi ortam sıcaklığı ile (t : °C) değişmektedir.

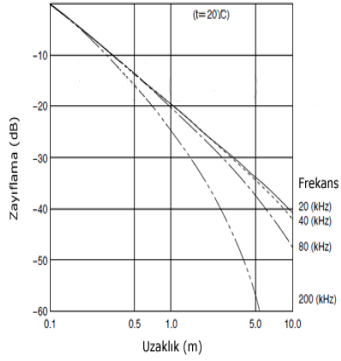

Şekil 5. Farklı frekanslardaki ses dalgalarının 20ºC'de hareket ve uzaklık eğrileri

Şekil 5 'te de görüldüğü gibi hava boşluğunda hareket eden ultrasonik dalgaların gücü uzaklığa oranla azalmaktadır. Yine grafikte görülebileceği üzere yüksek frekanslı ultrasonik dalgalar düşük frekanslı ultrasonik dalgalara oranla daha kısa mesafeye ulaşırlar.

*B. IP Kamera*

IP Kamera, üzerinde bulunan sensörler sayesinde görüntüyü yakalayıp bu görüntüleri dijital veriye çevirdikten sonra ağ (network) yardımıyla yetkilendirilmiş kullanıcılara ileten görüntüleme sistemidir.
IP kameraların genel özellikleri aşağıda sıralanmıştır.

- Çalışması için kart, kayıt cihazı vb. çeşitli aparatlara gerek duymayan ileri kamera teknolojisidir.
- İnternet ve network üzerinden erişim sağlanabilir.
- IP kameralar üzerinden gelen görüntü ve ses anlık olarak masaüstü / dizüstü bilgisayar, tablet ve mobil cihazlardan izlenebilmektedir.
- IP kameralar, programlanabilir bir yapıya sahip olduklarından görüntü depolama, hareket algılama, alarm yönetimi (hırsız, yangın vb.) gibi fonksiyonlara sahiptir.

IP kameraların çalışma sistemi, temel olarak yakaladığı görüntüyü kullanıcıya anlık olarak iletmektir. Şekil 6'da genel çalışma yapısı gösterilmiştir. İletimde öncelikle IP kameranın lens ve sensörlerinde oluşan görüntü elektrik sinyaline çevrilir. Bu elektrik sinyali, IP kameranın içinde dahili olarak bulunan işlemciye iletilir. İşlemci analog olan bu elektrik sinyalini dijital sinyale dönüştürür. Dijital sinyal, kayıt altına alınabileceği gibi kameraya dahili veya harici olarak bağlı network, ftp server, web server, smtp client üzerinden yayınlanması da mümkündür.

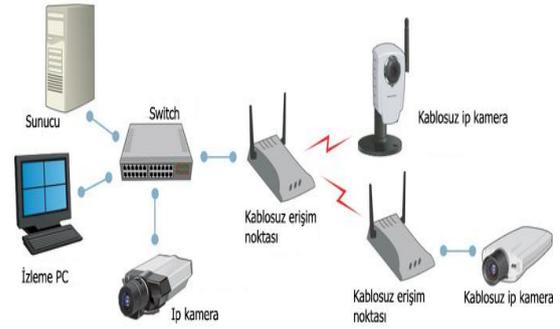

Şekil 6. IP kameranın çalışma şekli

IP Kameralar isminden de anlaşıldığı gibi Internet Protocol Kamera anlamında açılıma sahiptir. Bu açılım IP Kameraların bünyesinde bir Web Server bulundurmasını gerektirir. Cihazlar mevcut network' te sabit veya otomatik olarak atanacak bir IP numarasına ihtiyaç duymaktadırlar. IP ataması yapıldıktan sonra kablolu ya da kablosuz olarak bağlantı sağlanabilmektedir.

### III. SONUÇLAR

Günümüzde robotların kullanımının arttığı ve cep telefonu kullanımının çok yaygınlaştığı göz önüne alınarak, tasarlanan robot tamamen kablosuz olarak Android sistemli bir cep telefonundan bulunduğumuz ortamdan daha uzak bir ortamla iletişime geçmemizi ve oradan cep telefonuna anlık görüntüler ve çeşitli bilgiler göndererek oradaki ortam hakkında bilgi edinmemizi sağlamaktadır. Bu sayede hareket edemeyen, felçli, engelli, yatağa bağımlı hastalara refakat edebilmektedir. Refakatçi mobil olarak IP kamera ile hastayı kontrol edebilmekte ve hasta ile ilgili bilgi sahibi olabilmektedir. Böylece başka ortamlara gitmeden robotu kontrol ederek o ortamla iletişime geçmemizi sağlamaktadır. Refakatçinin hastanın yanında 24 saat kalamayacağı düşünülürse tasarlanan robotun bu konuda çok faydalı olabileceği görülmektedir.

Bununla beraber anne ve babaları çalıştığı durumlarda çocuklara mecburen bakıcı bakmaktadır. Bu durum aileler için oldukça tereddütlü bir durumdur. Bakıcıların olumsuz tutumlarına maruz kalan çocuklar sıklıkla medyadan izlenmektedir. Sabit kamera ile evin kontrolü oldukça yaygındır ve caydırıcıdır fakat kör noktalarda nelerin olduğu bilinmemektedir. Tasarlanan robot istenilen noktaya gidebildiği, IP kamerası hareketli olduğundan istenilirse bakıcı kontrol robotu olarak da kullanılabilir.

Bakıma ihtiyacı olan hastalar için nabız ve sıcaklık bilgisi kullanıcı tarafından kontrol edilebilmekte ve ayrıca anormal bir durum tespit edilmesi durumunda kullanıcıya sistem üzerinden uyarı bildirimi yapılmaktadır. Hastanın yanında bulunan acil durum butonu sayesinde herhangi bir acil durumda hastanın bu butonu kullanması ile sistem üzerinden kullanıcıya bilgilendirme mesajı gönderilmektedir. Android uygulamayla belirlenen ilaç kullanım çizelgesine göre ilaç kullanım zamanı geldiğinde hastaya sesli hatırlatma yapmaktadır. Otonom mod için robotun tekerlerine entegre edilmiş encoder yardımıyla alınan değerler kullanılarak geliştirilen algoritma ile robotun sürekli belirli noktalarda kalmasını engellemekte ve farklı noktalara hareket etmesini sağlamaktadır.

Ayrıca proje geliştirilerek, robotun hareket ağı genişletilip daha da uzak ortamlara gidebilmesi, yaşlı kimsenin ani düşmesi gibi bazı özel ve acil durumların makine öğrenmesi teknikleri ile robota öğretilmesi hedeflenmektedir.

## KAYNAKLAR

[1] D. Iñak, G. Jorge Juan, and S. Emilio, "Lower-Limb Robotic Rehabilitation: Literature Review and Challenges," Journal of Robotics, vol. 2011, Article ID 759764, 11 pages,

[2] F. S. David, and M. J. Maja. "Defining socially assistive robotics."9th International Conference on Rehabilitation Robotics, 2005. ICORR 2005 IEEE, 2005. , Chicago, IL, USA.

[3] T. S. Dahl, and M. N. K. Boulos. "Robots in Health and Social Care: A Complementary Technology to Home Care and Telehealthcare", Robotics 3.1 (2013): 1-21.

[4] Ö. Pektaş, "Mobil Bomba Robotlarının İncelenmesi ve Prototip Robot Tasarımı", İstanbul Üniversitesi, İstanbul 2010.

[5] Türkiye İstatistik Kurumu web sayfa: www.tuik.gov.tr

[6] S. Wild, G. Roglic, A. Green, R. Sicree, H. King, "Global prevalence of diabetes". Diabetes Care 27, 1047–1053, 2004.

[7] R. Looije, A. Mark Neerincx, and C. Fokie. "Persuasive robotic assistant for health self-management of older adults: Design and evaluation of social behaviors." International Journal of Human-Computer Studies 68.6 (2010): 386-397.

[8] T. Uzun, G. T. Erdoğan, "Bir Gezgin Robot için Elektronik Denetim Donanımının Tasarımı ve Uygulaması", Yıldız Teknik Üniversitesi, İstanbul 2011.

[9] İ. Şimşek, E. Durukan, F. E. Erdem, "Kroki Çizen Gezgin Robot", Karadeniz Teknik Üniversitesi, 2015.

[10] G. Gedikli, A. Temur, "Kablosuz Gamepad ile Araç kontrolü", Karadeniz Teknik Üniversitesi, Trabzon 2014.

[11] J. Borenstein, Y. Koren, "Obstacle Avoidance with Ultrasonic Sensors", IEEE Journal of Robotics and Automation, Vol. 4, No. 2, April 1988.

[12] O. Dongyue, H. Yuanhang, and Z. Yuting, "The Investigation of the Obstacle Avoidance for Mobile Robot Based on the Multi Sensor Information Fusion Technology", International Journal of Materials, Mechanics and Manufacturing, Vol. 1, No. 4, November 2013.

[13] A. K. Shrivastava, A. Verma, and S. P. Singh, "Distance Measurement of an Object or Obstacle by Ultrasound Sensors using P89C51RD2", International Journal of Computer Theory and Engineering, Vol. 2, No. 1 February, 2010.